\documentclass[letterpaper]{article} % DO NOT CHANGE THIS
\usepackage{aaai25}  % DO NOT CHANGE THIS
\usepackage{times}  % DO NOT CHANGE THIS
\usepackage{helvet}  % DO NOT CHANGE THIS
\usepackage{courier}  % DO NOT CHANGE THIS
\usepackage[hyphens]{url}  % DO NOT CHANGE THIS
\usepackage{graphicx} % DO NOT CHANGE THIS
\usepackage{natbib}  % DO NOT CHANGE THIS AND DO NOT ADD ANY OPTIONS TO IT
\usepackage{caption} % DO NOT CHANGE THIS AND DO NOT ADD ANY OPTIONS TO IT
\usepackage{algorithm}
\usepackage{algorithmic}

\usepackage{newfloat}
\usepackage{listings}
\DeclareCaptionStyle{ruled}{labelfont=normalfont,labelsep=colon,strut=off} % DO NOT CHANGE THIS
\floatstyle{ruled}
\newfloat{listing}{tb}{lst}{}
\floatname{listing}{Listing}
\usepackage{xspace}
\usepackage{amsmath}
\usepackage{adjustbox}
\usepackage{multirow}
\usepackage{booktabs}
\usepackage{xcolor}
\usepackage{todonotes}
\newcommand{\answerTODO}[1]{~(yes)}
\newcommand{\data}{EventSum\xspace}

\title{\data: A Large-Scale Event-Centric Summarization Dataset \\ for Chinese Multi-News Documents}
\author {
    Mengna Zhu\textsuperscript{\rm 1},
    Kaisheng Zeng\textsuperscript{\rm 2,3},
    Mao Wang\textsuperscript{\rm 1},
    Kaiming Xiao\textsuperscript{\rm 1},\\
    Lei Hou\textsuperscript{\rm 2}\footnotemark[1],
    Hongbin Huang\textsuperscript{\rm 1}\thanks{Corresponding Authors.},
    Juanzi Li\textsuperscript{\rm 2}
}
\affiliations {
    \textsuperscript{\rm 1}Laboratory for Big Data and Decision, National University of Defense Technology\\
    \textsuperscript{\rm 2}Department of Computer Science and Technology, Tsinghua University\\
    \textsuperscript{\rm 3}College of Information and Communication, National University of Defense Technology\\
    zhumengna16@nudt.edu.cn
}

\usepackage{bibentry}
\begin{document}
\maketitle

\begin{abstract}

In real life, many dynamic events, such as major disasters and large-scale sports events, evolve continuously over time. Obtaining an overview of these events can help people quickly understand the situation and respond more effectively. This is challenging because the key information of the event is often scattered across multiple documents, involving complex event knowledge understanding and reasoning, which is under-explored in previous work. 
Therefore, we proposed the \textbf{E}vent-\textbf{C}entric Multi-Document \textbf{S}ummarization (\textbf{ECS}) task, which aims to generate concise and comprehensive summaries of a given event based on multiple related news documents. Based on this, we constructed the \textbf{\data} dataset, which was constructed using Baidu Baike entries and underwent extensive human annotation, to facilitate relevant research. It is the first large-scale Chinese multi-document summarization dataset, containing 5,100 events and a total of 57,984 news documents, with an average of 11.4 input news documents and 13,471 characters per event. To ensure data quality and mitigate potential data leakage, we adopted a multi-stage annotation approach for manually labeling the test set. Given the complexity of event-related information, existing metrics struggle to comprehensively assess the quality of generated summaries. We designed specific metrics including Event Recall, Argument Recall, Causal Recall, and Temporal Recall along with corresponding calculation methods for evaluation. We conducted comprehensive experiments on \data to evaluate the performance of advanced long-context Large Language Models (LLMs) on this task. 
Our experimental results indicate that: 1) The event-centric multi-document summarization task remains challenging for existing long-context LLMs; 2) The recall metrics we designed are crucial for evaluating the comprehensiveness of the summary information. Our code and data can be obtained from \url{https://github.com/Mzzzhu/EventSum}.

\end{abstract}

\section{Introduction}
Dynamic events characterized by continuous development and change over time, uncertainty, and intricate causal relationships are pervasive in real life, such as natural disasters (\textit{earthquakes, floods}), major sports events (\textit{Olympics, UEFA European Championship}), and pressing social issues (\textit{criminal cases, sudden public health crises}), etc. 
These events are often covered by multiple news articles that report from different perspectives and may include real-time updates.
Integrating these diverse news sources is essential for a comprehensive understanding of the event.
Extracting key information from related news articles to create accurate and comprehensive summaries is crucial for quickly organizing information about the event and better supporting downstream applications such as opinion mining, intelligent assistants, emergency response, etc~\citep{opinion-mining, xu-etal-2020-xiaomingbot, emergency-res, emergency, urologin2018sentiment}.

\begin{figure}[t]
\centering
\includegraphics[width=1.0\columnwidth]{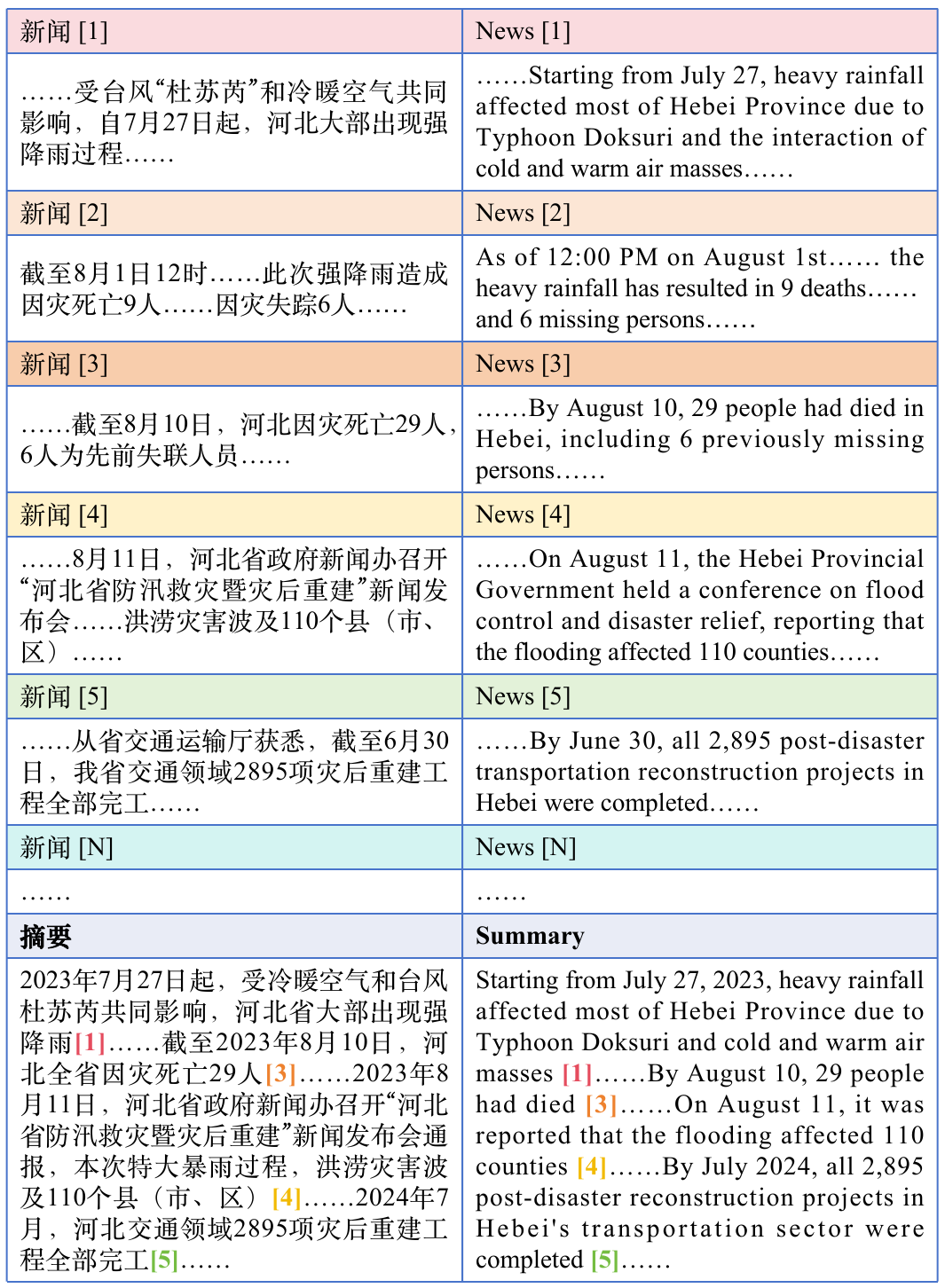} % Reduce the figure size so that it is slightly narrower than the column. Don't use precise values for figure width.This setup will avoid overfull boxes.
\caption{An example of the ``2023 Hebei Heavy Rain''. The right side of the figure shows the English translation obtained from the original documents on the left.}
\label{fig: mds-example}
\end{figure}

As illustrated in Figure~\ref{fig: mds-example}, the news articles provide information about the ``2023 Hebei Heavy Rain'': the cause (News 1: Typhoon Doksuri and cold and warm air), casualties (News 3: 29 deaths), affected areas (News 4: 110 counties), and measures taken (News 5: all post-disaster reconstruction projects completed).
 According to the generated summary which combined the information from the above news articles, the reader can quickly and conveniently grasp the full scope of the ``2023 Hebei Heavy Rain'' event.

Generating concise and comprehensive summaries based on multiple documents surrounding the specified event not only makes the content more comprehensible to humans but also offers richer information. However, most of current research on event understanding is based on single-document and structural comprehension~\citep{peng2023omnievent, wang2022maven, liu2020event, wang2020maven} and existing Multi-document Summarization (MDS) research faces three primary challenges: 1) News-focused datasets like Multi-News~\citep{multi-news} consist of news-related articles and corresponding summaries that are organized around general news content rather than the specific dynamic event in chronological order; 2) Most large-scale datasets are automatically constructed, like WikiSum~\citep{liu2018generatewikisum}, which compromises dataset quality and increases the risk of data leakage in the test set; 3) Common evaluation metrics such as the ROUGE for lexical evaluation~\citep{lin-2004-rouge} and BERTScore~\citep{bert-score} for semantic evaluation, are insufficient to adequately assess the completeness and comprehensiveness of summaries that focus on dynamic events.

To address the above challenges, in this paper, we constructed the first large-scale Chinese multi-news summarization dataset focused on dynamic events used for ECS, named \data. This dataset comprises a total of 5,100 events and 57,984 news articles, with each event corresponding to one piece of data. On average, each event has 11.4 related news articles and 13,471 input characters. In order to ensure the quality of the data, and the possibility of data leakage caused by the pre-trained corpus, we have implemented a multi-stage annotation method to manually write summaries for the test set data. These summaries are organized in chronological order of sub-events and retain structured information obtained through annotation, including key sub-events, key event arguments, and causal relationships, which are crucial for comprehensive event understanding.\footnote{After publication, the corresponding AAAI link for our paper will be updated here.}

The powerful capabilities of large language models provide solutions for many tasks in Natural Language Processing (NLP), but still face many challenges~\citep{tool4expert, qin2024large}. We used our dataset \data to evaluate the performance of several advanced long-text LLMs on the ECS task. To better assess the quality of generated summaries and its effectiveness in organizing event information, we developed specific key elements recall metrics, including Event Recall, Argument Recall, Causal Recall, and Temporal Recall. 
We used existing structured event understanding datasets to train Natural Language Inference (NLI) models to compute these metrics by judging whether key event elements were entailed in the generated summary. 
This approach allows for a more comprehensive evaluation of the recall rate of structured information, as annotated in the dataset, thereby providing a more detailed measure of the quality of the generated summaries.
The experimental results show that: 1) The event-centric multi-document summarization task remains challenging for current long-text LLMs on \data; 2) The designed recall metrics are significant for evaluating the comprehensiveness of the generated summary.

Our contributions can be summarized as follows:

1. We proposed the event-centric multi-document summarization task which generated summaries around specified dynamic events based on given multiple related documents. This task is beneficial for quickly organizing event information but also challenging because it requires a deep understanding of long texts and complex event information.

2. We developed \textbf{\data}, the first large-scale Chinese multi-document summarization dataset, automatically constructed from Baidu Baike entries for this task study. Compared to existing news-related MDS datasets, \data features the highest number of input documents and the longest input length. To address data leakage and ensure robust evaluation, we \textbf{manually wrote} summaries for the test set and annotated key event-related information.

3. We conducted comprehensive experiments using annotated data of \data to evaluate the performance of advanced long-text LLMs on this task. Given the complexity of event data, we designed specific recall metrics to better assess the generated summaries. Our experimental results highlight the challenges posed by this task and dataset while confirming the effectiveness of our designed metrics.

\section{Dataset Construction}
\begin{figure*}[ht]
\centering
\includegraphics[width=1.0\textwidth]{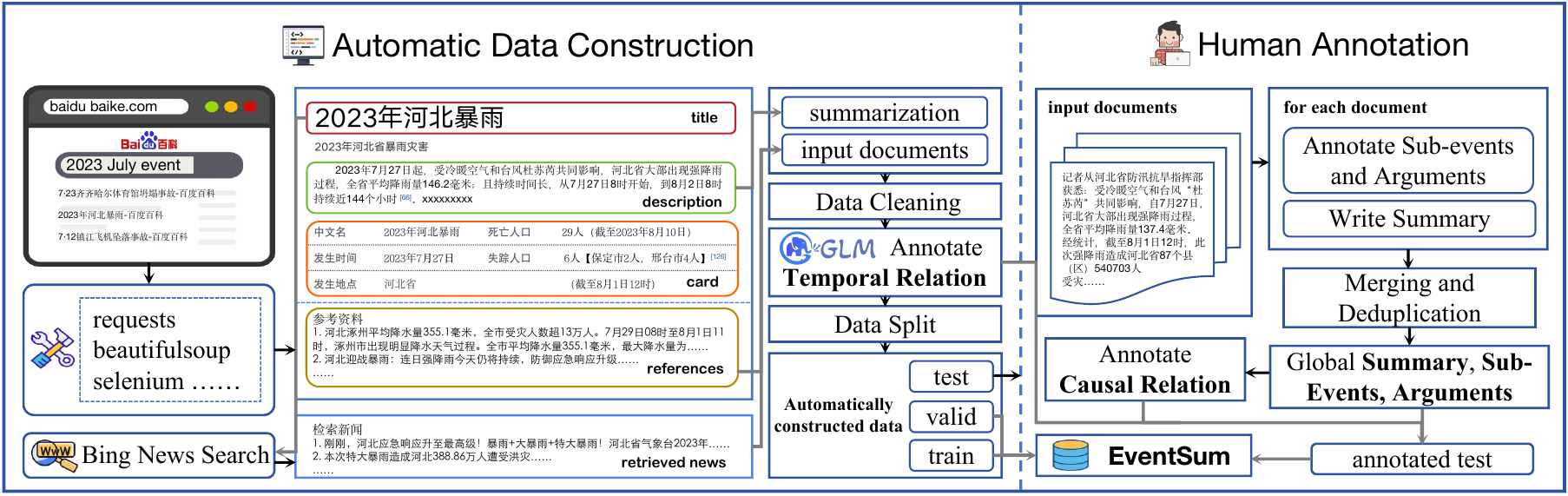}
\caption{Overview of the construction process. It introduces the data construction process for the ``2023 Hebei Heavy Rain'' event from the retrieved entries for ``Events of July 2023'' on the Baidu Baike website.}
\label{fig: contruction}
\end{figure*}

In this section, we provide a detailed introduction to the construction methods of the \data. The overview of our method can be illustrated in Figure~\ref{fig: contruction}, which shows an example of the data construction process for the entry ``2023 Hebei Heavy Rain''. It includes two parts: the automatic data construction process and the human annotation process.

\subsection{Automatic Data Construction}
The data for \data is sourced from event-related entries on Baidu Baike.\footnote{https://baike.baidu.com} The description information from these entries are used as reference summaries. 
Due to the potential absence of references or the omission of critical references in these entries, the input sources for the summaries are derived from two components: 1) News articles that correspond to the references listed within the entry, and 2) News articles retrieved based on the title information of the entry. 
The primary sources of these input news articles are reputable official news websites such as CCTV News, Huanqiu, and Sina\footnote{https://www.cctv.com; https://www.huanqiu.com; https://ww w.sina.com.cn}, which ensures the reliability of the information. The detailed construction process is as follows.

\paragraph{Data Collecting} We initially employed web scraping tools such as Requests, BeautifulSoup, and Selenium to harvest entries related to notable events from Baidu Baike between January 2000 and April 2024. We collected and stored the ``title'', ``card'', ``description'', and ``reference'' for each entry. Non-event entries were filtered out based on key fields such as ``time'' and ``location'' in the basic information table, resulting in a total of 14,000 entries.
To ensure the comprehensiveness of the summary information sources, we utilized the Bing News Search API\footnote{https://www.microsoft.com/en-us/bing/apis/bing-news-search-api} to retrieve related news articles for each entry based on its title. We specifically selected 20 news articles published within a month of the event date based on the ``time'' field in the card which contains basic event information as supplementary input documents. 

\paragraph{Data Cleaning} We cleaned and filtered the input documents through techniques like regular expression matching, removing some missing and duplicate documents. To further minimize noise introduced during document retrieval, we utilized the sentence-transformers\footnote{The model we used is paraphrase-multilingual-mpnet-base-v2.}~\citep{reimers-2020-multilingual-sentence-bert} to calculate the textual similarity between the retrieved documents and the summaries, filtering out low-relevance documents with a similarity score below the preset threshold of $0.5$ and the number of input documents was controlled between $5$ and $20$. 

\paragraph{Temporal Relation Annotation} Considering that temporal relationships are relatively simpler compared to other types of event information, and in the summaries generated around dynamic events, the events we are concerned with that have temporal relationships usually appear with clear time information indicators or obvious conjunctions. Therefore, we used LLMs\footnote{LLMs used here and in Designed Recall Metrics are glm-4-9b.} to automatically annotate temporal relationships in the reference summaries, and only annotated event pairs with clear ``before'' and ``after'' relationships. Considering the transitivity of temporal information, we only label events that are directly adjacent in time. We obtained 14,932 temporal relationships in total.

Through the aforementioned steps, we automatically construct data pairs, which can be represented as $i = [D, r, T] $, where $D$ represents the set of input documents, $r$ represents the reference summary, and $T$ represents the set of temporal relationships automatically annotated based on $r$. 

Finally, we obtain 5,100 instances and split them into training, validation, and testing sets. In \data, each instance corresponds to a dynamic event.

\subsection{Human Annotation}
Considering the efficiency and cost of manual summarization, we chose to manually annotate the test set of EventSum due to the inherent data leakage issues associated with open-source data. To guarantee the data quality, we adopted a multi-stage annotation method, replacing the original reference summary $r$ in the test set with the manually written summary $r'$. Detailed annotation process is as follows.

\paragraph{Sub-events and Arguments Annotation} Annotate structural sub-events and their relevant argument information related to the core dynamic event in each input document. The sub-events were annotated as \textbf{sentences} containing key event information, and the arguments we focused on annotating included ``time'', ``location'', ``person'', and ``organization''. The definitions of the event and arguments mentioned above are widely adopted, similar to those in ACE 2005~\citep{ace2005}. However, it is important to note that our work does not define any specific event schema.

\paragraph{Summary Writing} Write a summary for each input document. During the summarization process, should consider the structured event information annotated in Step 1 and use expressions from the input documents as much as possible.

\paragraph{Global Information Annotation} Deduplicate structured event information from the annotations of each document and organize the summaries chronologically to generate a global summary and compile structured event information.

\paragraph{Causal Relation Annotation} Annotate causal relationships between sentences in the global annotated sub-events. Following MAVEN-ERE~\citep{wang2022maven}, the annotated causal relationships primarily include ``\textit{cause}'' and ``\textit{precondition}'', where ``\textit{cause}'' indicates a sufficient condition, and ``\textit{precondition}'' indicates a necessary condition.

Through this multi-stage annotation method, we ensure that the annotated summaries contain more comprehensive, complete, and accurate event-related information. The annotated data is iteratively checked and corrected to ensure high annotation quality.  In the annotated data, there are 2,345 sub-events, 4,787 arguments, and 1,107 causal relationships.

Ultimately, each instance in the testing set is represented as $i = [D, r', T, G]$, where $r'$ represents the newly manually written reference summary, and $G$ represents the manually annotated global structured information.

\subsubsection{Quality Control}
To ensure data quality, annotators were divided into three groups with cross-checks during annotation. Project managers conducted random checks and resolved conflicts, while acceptance reviewers verified manager-approved data, calculated pass rates, and provided revision requirements. This process continued until the data achieved a 90\% pass rate. The pass rate is calculated as (number of data meeting the criteria / total data) * 100\%. Criteria: 1) documents are relevant to the event, and 2) the summary covers key event elements and organized in chronological order. We sampled 50 instances from both automatically constructed and human-annotated data for review. The pass rate for automatic data was 81\%, and for human-annotated data was 93\%. For temporal relationships annotated by LLMs, we reviewed samples to ensure proper labeling of sub-events with clear time indicators or conjunctions. When the labeled main temporal relationships reach 80\%, the data point is qualified. The qualified rate is 83\%.

\subsection{Data Analysis}
The final constructed \data dataset contains 5,100 instances and 57,984 news articles. We choose to compare \data with Multi-News~\citep{multi-news}, DUC data from 2003 and 2004~\cite{over2004introduction-DUC2004}, and TAC 2011~\cite{tac-2011} data, which are typically used in multi-document settings and focused on news. The comparison result is shown in Table~\ref{tab:dataset_comparison}. It can be observed that \data is the first Chinese dataset specifically designed for multi-document summarization (MDS). Moreover, the number of input characters far exceeds that of other datasets. The number of input documents is also significantly higher compared to the widely used Multi-News.

\begin{table*}[t!]
    \centering
    % \begin{adjustbox}{max width=1\linewidth}
{
    \begin{tabular}{cccccc}
    \toprule
    {\textbf{Dataset}} & \textbf{Language} & \textbf{Data Size} & \textbf{\#Docs} & {\textbf{\#Words (Input)}} & \textbf{\#Words (Output)}\\
    \midrule
    DUC 03+04 & English & $320$ & $10$ & $4,636$ & $110$\\
    TAC 2011 & English & $176$ & $10$ & $4,696$ & $100$\\
    Multi-News & English & $44,972/5,622/5,622$ & $2.7$ & $2,104$ & $264$\\
    \textbf{\data} & Chinese & $4,015/500/585$ & $11.4$ & $13,471$ & $161$ \\
    \bottomrule
    \end{tabular}
}
    % \end{adjustbox}
    \caption{Comparison between \data and existing MDS datasets with existing datasets that are focused on news and most similar to our data. Words represent tokens for English datasets and characters for Chinese datasets.}
    \label{tab:dataset_comparison}
\end{table*}

To better understand the characteristics of \data, we conducted a detailed statistical analysis on the input news documents and the reference summary as follows.

\begin{figure}[t]
\centering
\includegraphics[width=1.0\columnwidth]{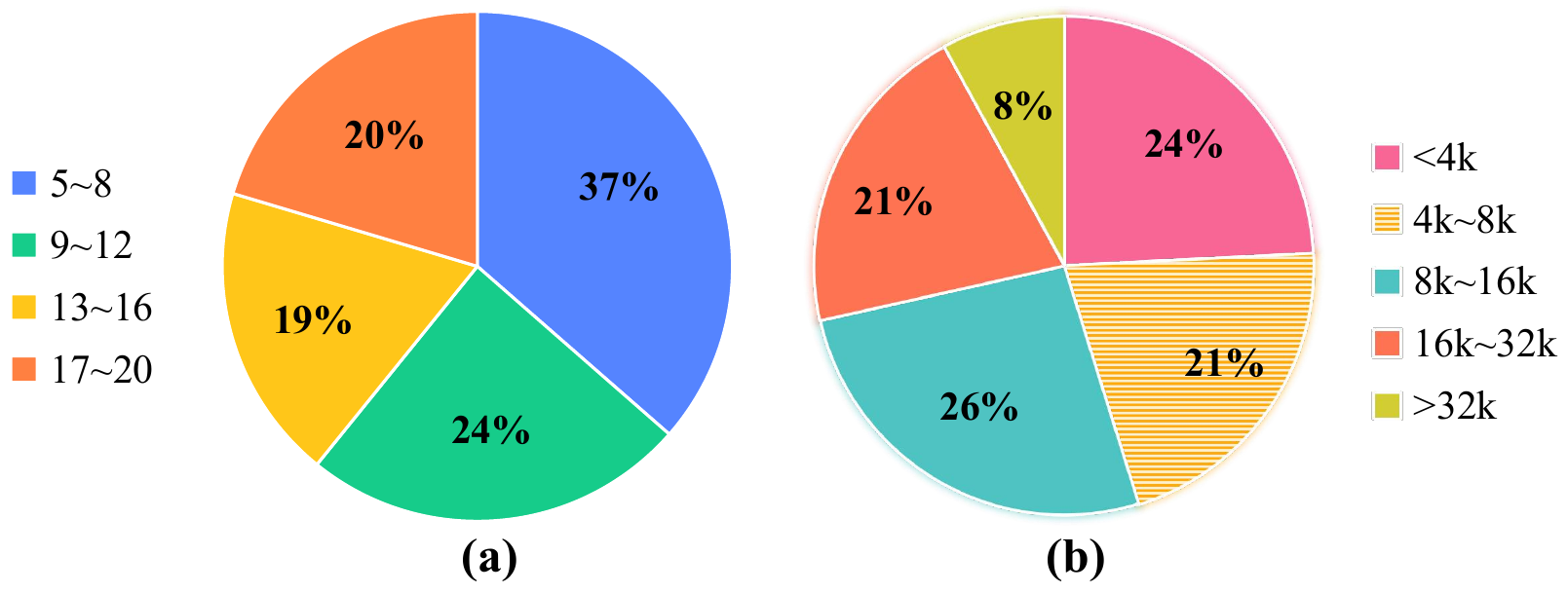}
\caption{Analysis of Input Documents. The distribution of the number of input documents (a) and the total input characters (b) are presented on the left and right, respectively.}
\label{fig: docs_dist}
\end{figure}

\paragraph{Analysis of the Input Documents}
The number of input documents was controlled between 5 and 20. The average number of input documents is 11.4 and the distribution of the number of input documents can be seen in Figure~\ref{fig: docs_dist} (a). Most instances have more than 8 input documents and one-fifth of the instances have more than 16 input documents.

The average input length is 13,471 characters, with a maximum length of 174,152 characters. The distribution of input lengths is shown in Figure~\ref{fig: docs_dist} (b). Over half of the instances contain more than 8,000 characters in the input documents, and nearly one-third have more than 16,000 characters.

\begin{figure}[t]
\centering
\includegraphics[width=0.96\columnwidth]{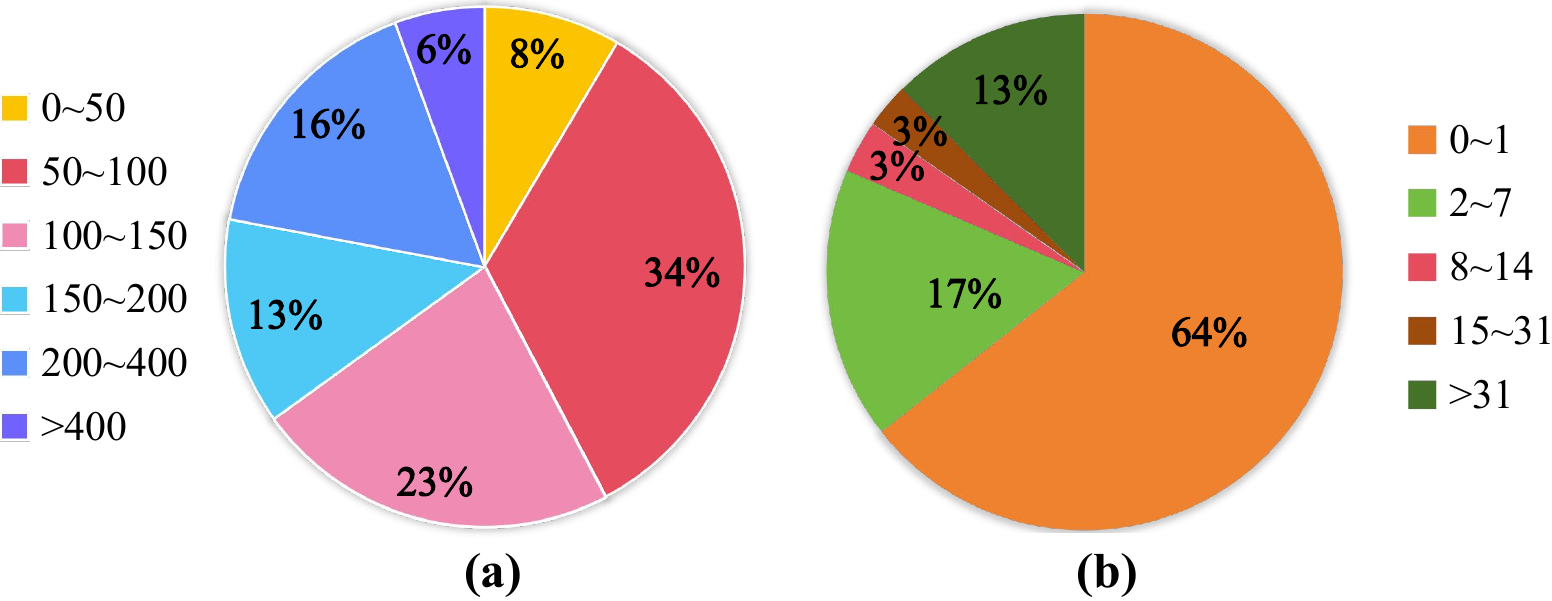}
\caption{Analysis of Reference Summary. The distribution of characters in the reference summary is shown on the left (a), while the distribution of the dynamic event time span within the reference summary is displayed on the right (b).}
\label{fig: sum_dist}
\end{figure}

\paragraph{Analysis of the Reference Summary}
We conducted an analysis of the length distribution of the reference summaries, as illustrated in Figure~\ref{fig: sum_dist} (a). The average length of the summaries is 161 characters, which meets the requirement for conciseness in practical applications. Additionally, we also analyzed the time span corresponding to the dynamic event in the reference summaries, as shown in Figure~\ref{fig: sum_dist} (b). Nearly 40\% of the data spans more than one day, and 13\% spans more than one month, reflecting the distribution of events in real-world scenarios.

\section{Evaluation Metrics}
\label{sec: metrics}
Evaluating metrics are essential for measuring the quality of the generated summaries in the summarization task, and well-defined metrics are crucial for relevant research~\citep{mds-survey}. In this section, we introduce the evaluation metrics employed in our study in detail, including commonly used existing metrics and our specifically designed metrics.

\paragraph{Common Metrics} ROUGE is the most commonly used metric in the summarization community and it comprises a set of evaluation metrics that assesses the similarity between the generated summary $s$ and the reference summary $r$. It includes multiple variants to evaluate candidate summaries in different ways, with the most commonly used being ROUGE-N and ROUGE-L. ROUGE-N measures the n-gram recall between $s$ and $r$. ROUGE-1 and ROUGE-2 are special cases of ROUGE-N that are usually chosen as best practices and represent the unigram and bigram, respectively. ROUGE-L adopts the longest common subsequence algorithm to count the longest matching vocabularies.

BERTScore is a prominent semantic matching metric that leverages pre-trained BERT embeddings to compute the similarity between the tokens in the generated summary $s$ and the reference summary $r$. This approach provides a context-aware measure of semantic equivalence.

In our paper, we used F1-scores of ROUGE-1, ROUGE-2, ROUGE-L and BERTScore for evaluation.

\paragraph{Designed Recall Metrics} Given that our task focuses on event-centric summarization, in order to better evaluate the completeness and accuracy of event information in the generated summaries, we specifically designed key element recall metrics, including \textit{Event Recall}, \textit{Argument Recall}, \textit{Causal Recall} and \textit{Temporal Recall}.

Due to the diversity of text generation, it is not feasible to directly calculate the recall rate of key elements using simple methods like regular expression matching. Drawing inspiration from the Recognising Textual Entailment task which defines textual entailment as that one text fragment can be inferred from another text fragment~\citep{dagan-texttual-entailment}, we obtain relevant recall rate by judging whether key elements were entailed in the generated summary or not. The general formula for key elements is:

\begin{equation}
\operatorname{Recall}_{k_i} = \frac{\sum_{e \in \mathcal{E}_{k_i}} \Gamma(e, s)} {\vert \mathcal{E}_{k_i} \vert},
\label{eq: recall}
\end{equation}

\begin{equation}
\Gamma(e, s) =
\begin{cases} 
    1, & if\ e\ \subseteq \ s,\\
    0, & else.
\end{cases}
\label{eq: entail}
\end{equation}
Here, $e$ represents the key element annotated based on the reference summary, $\mathcal{E}_{k_i}$ denotes the set of relevant annotated key elements with type ${k_i}$, and $s$ denotes the generated summary. $\Gamma(e,s)$ is a discriminator used to determine whether the annotated element $e$ can be inferred from the generated summary $s$ or not. 
The entailment is confirmed only when $\Gamma(e,s)=1$, indicating that the element $e$ is indeed entailed in the summary $s$. The $\subseteq$ means the entailment relationship. 

In the equation~(\ref{eq: recall}) and~(\ref{eq: entail}), the corresponding $e$ to designed metrics \textit{Event Recall}, \textit{Argument Recall}, \textit{Causal Recall} and \textit{Temporal Recall} is the sentence containing key event information, event-related arguments, causal relationships, and temporal relationships respectively.

We used existing event understanding datasets and the automatically constructed data of \data to train a relevant binary classification Natural Language Inference (NLI) model as the discriminator. Specifically, CMNEE~\citep{zhu2024cmnee} is used to construct data for training NLI models for Event Recall and Argument Recall as it is a large-scale Chinese event extraction dataset and annotates coref-arguments information such as abbreviations, pronouns, etc. Considering there is currently no suitable Chinese dataset for event causal relationships extraction study and causal relationships are relatively more complex, making it difficult to obtain reliable annotation automatically, we chose to use the translated version of MAVEN-ERE~\citep{wang2022maven} to construct data for training the NLI model for Causal Recall because the causal relation annotation requirement is similar to that of \data.
Additionally, the automatically constructed data of \data, where temporal relationships were annotated during the dataset construction process, was used to train the NLI model for Temporal Recall.

The structural annotation information is converted into natural language expression by LLMs as $t_2$ for positive instances, with input text designated as $t_1$ for the NLI models. To better evaluate the quality of the summaries, we analyzed the generated summaries and designed three strategies to construct negative instances by modifying a certain proportion of the positive instances, making the constructed data more closely aligned with our actual requirements.
The negative instances generation strategies are as follows.
\begin{itemize}
\item \textbf{Remove}: Use the sentence-transformers library to evaluate the similarity between sentences in $t_1$ and $t_2$. Sentences from $t_1$ with a similarity score exceeding a threshold of 0.5 should be removed. The remaining sentences are then concatenated to form a new text, denoted as $t_1'$.
\item \textbf{Revise}: Instruct the LLM to modify key event-related information in $t_2$, such as ``time'', ``location'', ``quantity'' and ``person'', etc. Alternatively, the LLM may expand upon or remove certain details surrounding the key event in $t_2$. The modified sentence is then used as $t_2'$.
\item \textbf{Replace}: Randomly retrieve 100 instances, calculate the similarity between the text of the retrieved instances and $t_2$, and use the text with the highest similarity but no overlapping event information as the new text $t_1'$.
\end{itemize}

Then we trained models to determine whether $t_2$ ($t_2'$) was entailed in $t_1$ ($t_1'$) or not. The model we selected to train was chinese-roberta-wwm-ext~\citep{chinese-bert-wwm} which provides robust support for the Chinese corpus\footnote{More details about our trained NLI models can be found from appendix B: Details of NLI models.}.

On the test set of our constructed data for NLI models, final $Event\ Recall$, $Argument\ Recall$, $Causal\ Recall$, and $Temporal\ Recall$ are $96.8$, $92.9$, $94.3$, $92.1$ respectively.

We used trained models as discriminators to determine whether key elements annotated in the test set of \data were entailed in the generated summaries. The generated summary is as ``$t_1$'' and the key element is as ``$t_2$''. In this way, we could obtain all the recall metrics we need.

\section{Experiments}
In this section, we used the annotated data of \data to evaluate the performance of advanced long-context LLMs on the event-centric multi-documents summarization task.

\subsection{Experimental Setup}
We evaluate 10 popular LLMs that feature long context capability and good support for Chinese, including open-source models: Baichuan2-13b-chat~\citep{baichuan2023baichuan2}, Llama3-chinese-8B-Instruct~\citep{chinese-llama-alpaca}, Yi-1.5-9b-chat-16k~\citep{ai2024yi}, Qwen2-7b-Instruct~\citep{qwen2}, glm-4-9b-chat~\citep{glm2024chatglm}, InternLM2.5-7B-Chat-1M~\citep{cai2024internlm2}, glm-4-9b-chat-1M~\citep{glm2024chatglm}, and commercial models: MoonShot~\citep{qin2024mooncake}, Claude-3-Opus~\citep{anthropic2024claude}, GPT-4o\footnote{The models we used: moonshot-v1-128k, Claude-3-Opus-20240229, GPT-4o}~\citep{openai2024gpt4technicalreport}. Metrics we used have been introduced above. The assessment was conducted under the zero-shot setting\footnote{More details can be found in the appendix C: Experimental Details. Few-shot experimental results can be seen in the appendix D: Few-shot Experimental Results.}.

\subsection{Overall Results}

 Overall experimental results indicated that the task and our dataset are challenging, as shown in Tabel~\ref{tab:overall_result},  We summarized our findings from following aspects:
 
 1) Performance on Commonly Used Metrics: Among the open-source models, the best model was \textit{glm-4-9b-1M}, while the best commercial model was \textit{GPT-4o}. There is still significant room for improvement in overall performance. Open-source models outperformed commercial models on commonly used metrics. This is mainly because these models have undergone targeted training in Chinese, resulting in more natural and fluent expression in our task.
 
 2) Effect of Input Length Limitation on Performance: Comparing the performance of open-source models with different input length limitations revealed that longer input length improved performance. This can be obviously seen from the results of \textit{glm-4-9b-chat} and \textit{glm-4-9b-chat-1M}.
 
 3) Performance on Event-Centric Metrics: We observed a trend opposite to that seen with common metrics. \textit{Claude-3-Opus} performed best on almost all our designed metrics, and commercial models generally performed better than most open-source models, which indicates the importance of our designed metrics for comprehensive evaluation.
 
 4) Analysis of Our Designed Metrics: Specifically analyzing our designed metrics, we found that Event Recall was significantly lower compared to other metrics. This is mainly because the expression of sub-events is more complex than that of arguments and there is a greater quantity of sub-event data compared to relationships data.

 \begin{table*}[t!]
    \centering
{
    \begin{tabular}{l|l|c|rrrr|rrrr}
    \toprule
    {\textbf{Type}} & {\textbf{Model}} & {\textbf{Length}}& {\textbf{R-1}} & {\textbf{R-2}} & {\textbf{R-L}} & {\textbf{BS}} & {\textbf{ER}} & {\textbf{AR}} & {\textbf{CR}} & {\textbf{TR}} \\
    \midrule
    \multirow{7}{*}{Open-Source} & 
      Baichuan2-13b-chat & $8$K & $30.3$ & $15.9$ & $22.7$ & $66.6$ & $15.7$ & $21.9$ & $23.3$& $18.6$ \\
    & Llama3-chinese-8B-Instruct & $8$K & $40.1$ & $22.8$ & $29.8$ & $73.5$ & $15.1$ & $26.9$ & $24.4$ &  $20.2$\\
    & Yi-1.5-9b-chat-16k & $16$K & $35.7$ & $18.2$ & $19.8$ & $68.0$ & $14.8$ & $31.4$ & $50.3$ & $39.6$ \\
    & Qwen2-7b-Instruct & $32$K & $47.2$ & $26.6$ & $32.1$ & $75.8$ & $\mathbf{27.4}$ & $\underline{48.4}$ & $\underline{66.7}$ & $\underline{53.5}$ \\
    & glm-4-9b-chat & $128$K & $\underline{48.2}$ & $\underline{27.0}$ & $\underline{35.6}$ & $\mathbf{77.2}$ & $16.2$ & $32.8$ & $22.8$ & $15.3$\\
    & InterLM2.5-7B-Chat-1M & $1$M & $47.9$ & $26.7$ & $33.7$ & $76.3$ & $24.5$ & $44.2$ & $55.2$ & $40.5$\\
    & glm-4-9b-chat-1M & $1$M & $\mathbf{49.3}$ & $\mathbf{28.6}$ & $\mathbf{36.2}$ & $\underline{77.0}$ & $23.9$ & $43.8$ & $47.5$ & $31.8$\\
    \midrule
    \multirow{3}{*}{Commericial} & MoonShot & $128$K & $43.2$ & $23.2$ & $29.9$ & $71.9$ & $23.5$ & $43.7$ & $57.9$ & $42.3$\\
    & Claude-3-Opus & $200$K & $45.1$ & $22.9$ & $29.7$ & $75.2$ & $\underline{25.7}$ & $\mathbf{50.3}$ & $\mathbf{67.3}$ & $\mathbf{56.8}$\\
    & GPT-4o & $128$K & $47.5$ & $26.1$ & $33.1$ & $76.2$ & $21.7$ & $46.2$ & $56.1$ & $40.0$\\
    \bottomrule
    \end{tabular}
    }
    \caption{Experimental results on \data. Length: Input length limitation of models; R-1: ROUGE-1; R-2: ROUGE-2; R-L: ROUGE-L; BS: BERTScore; ER: Event Recall; AR: Argument Recall; CR: Causal Recall; TR: Temporal Recall. Metric Definitions were illustrated in the Evaluation Metrics section. The best results are in bold. The second-best results are underlined.}
    \label{tab:overall_result}
\end{table*}

\section{Further Analysis}
In this section, we randomly sampled 50 instances from the prediction results of the best-performing model, \textit{Claude-3-Opus}, for manual observation to gain further insights into \data. Additionally, we analyzed the impact of the number of input documents and different time spans on performance, and assessed the reliability of the evaluation metrics to ensure a comprehensive analysis.

\subsection{Analysis of Generated Summaries}
To better understand the challenges of \data, we observed the generated summaries of the sampled data and summarized the common issues into 3 main categories~\footnote{Example cases can be seen in the appendix E: Cases for Generated Summaries Analysis }: 

1) Incomplete or Missing Information: The summaries might omit key elements of the dynamic event including sub-events, arguments, causal relations, and temporal relations mentioned above. This can lead to summaries lacking a comprehensive description of the dynamic event, as indicated by the results of our recall metrics.

2) Over or Under generalization: Summaries may be too vague, failing to capture specific details of the dynamic event, or too detailed, making the summary unnecessarily long. Striking the right balance between detail and brevity is a common challenge for the summarization task.

3) Irrelevance: Summaries might include irrelevant information that is not directly related to the dynamic event like reflective or commentary content or even other events information, especially when news like summary reports include multiple similar events appear in the input documents.

Additionally, in some models with poorer performance, issues such as repetition, incoherence, and poor responsiveness to the instructions in the prompt were also observed.

\subsection{Metric Evaluation}
Referring to the evaluation method of ROUGE, we compared the recall metrics obtained by trained NLI models with manually computed results based on human evaluation and also calculated their consistency on our sampled data to assess the reliability of our designed metrics, as shown in Table~\ref{tab:metric_result}. The results in the table are close and relevant consistency all exceeds 90\%. Most metrics from trained NLI models are generally slightly lower compared to human evaluation. This is because the high degree of diversity in the text generated by LLMs makes it difficult to identify some entailment relationships. However, this does not affect the comparison of the capability of LLMs, and the differences between the results are within an acceptable range. It indicates that our trained models can effectively compute the recall rate of key elements and better evaluate the quality and completeness of the summary.

\begin{table}[t!]
    \centering
    % \begin{adjustbox}{max width=1\columnwidth}
{
    \begin{tabular}{lccc}
    \toprule
    {\textbf{Metric}} & {\textbf{Predicted}} & \textbf{Human} & \textbf{Consistency}\\
    \midrule
    Event Recall & $19.2$ & $24.2$ & $95.0$ \\
    Argument Recall & $36.4$ & $39.4$ & $97.0$ \\
    Causal Recall & $67.7$ & $71.7$ & $90.9$\\
    Temporal Recall & $50.5$ & $49.5$ & $92.9$\\
    \bottomrule
    \end{tabular}
}
    % \end{adjustbox}
    \caption{Recall metrics obtained by trained NLI models and human judgment followed by consistency between them.}
    \label{tab:metric_result}
\end{table}

\subsection{Impact of Number of Input Documents}
The impact of the Number of Input Documents can be seen in Figure~\ref{fig: futher_analysis} (a). It shows that almost all metrics exhibit clear overall downward trends. However, the trends for temporal and causal relationships recall differ from the other metrics. The best performance is not achieved with the fewest input documents. This suggests that the model may better capture the relationships between events within a certain range of input document quantities, but as the number of input documents further increases, the complexity of the information rises, leading to a decline in the performances.

\subsection{Impact of Time Span}
\begin{figure}[t]
\centering
\includegraphics[width=1.0\columnwidth]{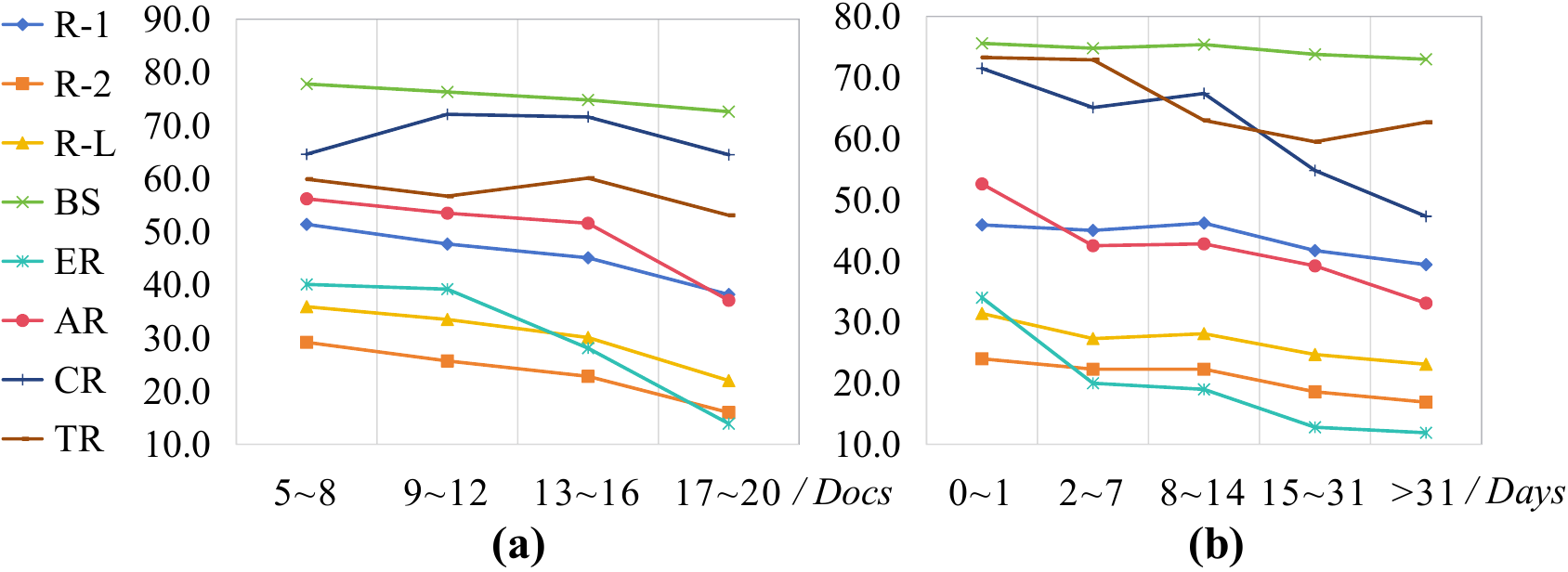}
\caption{Analysis of the impact of various input documents number (left) and time span (right) of the dynamic event.}
\label{fig: futher_analysis}
\end{figure}

The impact of time span of the dynamic event can be seen in Figure~\ref{fig: futher_analysis} (b). It can be observed that all metrics show an overall downward trend. The decline is more pronounced in our designed recall metrics compared to the common metrics. As the time span of the event increases, the number of corresponding sub-events, event arguments, causal relationships, and temporal relationships also increases, making the event information more complex and challenging to ensure the comprehensiveness and completeness of the generated summary. The change in the BERTScore, a metric for semantic matching, is relatively small, indicating that LLMs generally maintain good semantic relevance.

\section{Related Work}
Text summarization is an important research area in the field of NLP~\citep{tas2007survey, el2021automatic, automated-review}. Multi-document summarization (MDS), due to the need to integrate information from multiple sources related to a specific topic, is even more challenging. Our work can be seen as an important extension of MDS. In this section, we will introduce some representative datasets and evaluation metrics for MDS.

\paragraph{Datasets}
MDS prioritizes capturing key information across documents and emphasizes content coverage across documents without being constrained by specific temporal or event-based structures. Except for Multi-News, DUC and TAC datasets we compared with \data above that focused on News, there are also many other datasets for MDS, such as WikiSum~\citep{liu2018generatewikisum}, WCEP~\citep{wcep} constructed based on Wikipedia, Multi-XScience~\citep{lu2020multixsciencelargescaledatasetextreme} focused on scientific articles, GameWikiSum~\citep{antognini2020gamewikisum} focused on game vedio, Opinosis~\citep{ganesan2010opinosis}, Yelp~\citep{chu2019-yelp} focused on reviews, etc. In MDS, there is a specific task similar to ours, namely Timeline Summarization (TLS)~\citep{timeline, yan-etal-2011-timeline}, which requires generating daily summaries and arranging them in chronological order. The mostly used dataset for TLS is Timeline17~\citep{timeline17} and Crisis~\citep{crisis}. Timeline17 contains 17 topics and 19 timelines in total. Crisis has 5 topics and 22 timelines annotated in total. Available data for TLS is limited, which impedes relevant research. Our research can offer insights into how to construct TLS datasets and may even serve as a potential resource for TLS.

Considering ECS not only requires an understanding of temporal progression, like TLS, but also demands the ability to delineate the core event and sub-events, capture event relationships (co-reference, causal, etc.), resolve conflicting information, and integrate updates from multiple sources, which is helpful for obtaining a deep understanding of event dynamics while providing a clear and comprehensive view, existing MDS datasets are not well-suited for our task.

\paragraph{Evaluation Metrics} Conventional evaluation metrics used in the MDS research can be mainly divided into two categories: 1) lexical matching metrics which evaluate the similarity between generated summaries and reference summaries based on exact word overlaps, such as ROUGE~\citep{lin-2004-rouge}, BLEU~\citep{papineni2002bleu}, Pyramid~\citep{pyramid}; 2)semantic matching metrics which evaluate the meaning and contextual relevance of generated summaries beyond surface-level word overlaps, such as BERTScore~\citep{bert-score}, Moverscore~\citep{zhao2019moverscore}, METEOR~\citep{banerjee2005meteor}. These metrics have been shown to have a relatively low correlation with human judgments, especially for tasks with creativity and diversity~\citep{zhang2024systematicsurveytextsummarization, wang2023surveyinglandscapetextsummarization}. There are also some specialized metrics commonly used in TLS tasks. Except for the commonly used Rouge series scores including concat F1, Agree F1 and Align F1, Date F1 is mostly used for key date selection evaluation~\citep{timeline-event-graph, hu-etal-2024-moments}. With the development of LLMs and their outstanding performances in various natural language processing tasks, a series of recent work has tried to use LLMs for evaluation~\citep{wu2023llm-evaluate, liu-etal-2023-geval, llm-benchmarking}. 

To conclude, the existing evaluation metrics in MDS are not suitable for accurately evaluating the event content and cannot effectively assess the comprehensiveness of event information in the generated summaries.
 
\section{Conclusions and Future Work}
We proposed the first large-scale, event-centric summarization dataset for Chinese multi-news documents, \data, which was automatically constructed based on Baidu Baike entries, along with a manually annotated test set to mitigate the impact of inherent data leakage. Given the event-centric nature of \data, we designed recall metrics, including Event Recall, Argument Recall, Causal Recall, and Temporal Recall, to complement commonly used metrics in summarization tasks for evaluation. Experimental results demonstrated that this task and dataset are challenging, and further analysis confirmed the effectiveness and importance of our designed metrics. In the future, we plan to extend our approach to English corpora and increase the proportion of long time span events. Additionally, we hope to explore more sophisticated methods to conduct extensive experiments and further enhance performance.

\section{Ethical Statement}
This paper presents a new dataset, and we discuss some related ethical considerations here: (1) \textbf{Copyright Statement.} All data utilized in this work are publicly available and freely accessible, with no inclusion of proprietary or restricted data. The use of these datasets strictly adheres to the terms and conditions of their respective platforms. As such, this work does not involve any copyright infringement or related issues. (2) \textbf{Worker Treatments.} We hire annotators from a professional data annotation company, ensuring fair compensation with agreed-upon salaries and workloads. All employment terms are contractually bound and adhere to local regulations. (3) \textbf{Risk Control.} Given that the texts in our dataset \data do not contain private information and are sourced from open data, we believe \data poses no additional risks. To verify this, we manually reviewed a random sample of the data and found no concerning issues.

\section{Acknowledgments}
We thank all the anonymous reviewers and meta
reviewers for their valuable comments, as well as all of our team members for their support and assistance. This work is supported by Beijing Natural Science Foundation (L243006) and Natural Science Foundation of China (62476150).

\bibliography{aaai25}

\appendix
\section{A. More details about \data}
\label{more details of dataset}
\begin{figure}[t]
\centering
\includegraphics[width=0.95\columnwidth]{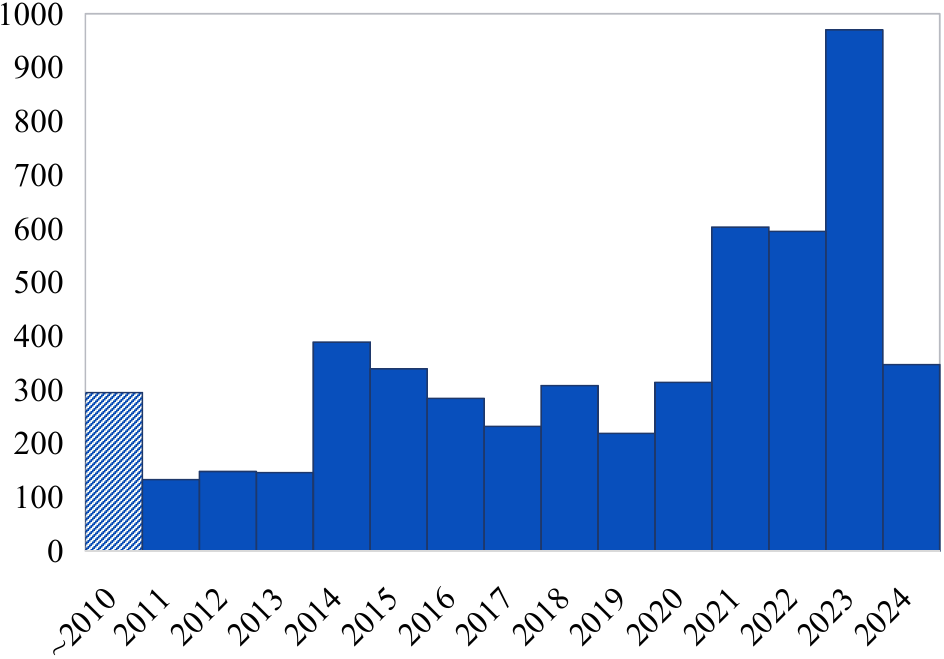}
\caption{Analysis of Event Occurrence Time Distribution. The year ranges from 2000 to 2024.}
\label{fig: year_dist}
\end{figure}

Event occurrence time ranges from the year 2000 to 2024 (until April, 2024) and detailed distribution can be seen in Figure~\ref{fig: year_dist}. Most of the events selected in the dataset occurred in the last 10 years (with 53.1\% of events occurring in the last 5 years). Therefore there is less possibility of input news missing, which helps to ensure the completeness and the quality of our data.

\section{B. Details of NLI models}
\label{appendix:NLI}
In this section, we introduce the detailed process for our trained NLI models.

\subsection{Data Preparation}
To train Natural Language Inference (NLI) models capable of determining whether key event-related information $e$ is entailed in the generated summary $s$, and to provide a more comprehensive evaluation of summary quality, we constructed $[t_1, t_2, label]$ data pairs using existing event understanding datasets as well as automatically constructed data from \data, which includes temporal relationship information.

In the constructed dataset, $t_1$ represents the input text, $t_2$ represents the key element information, and the label value falls within $[0,1]$. A label of ``1'' indicates that $t_2$ is entailed in $t_1$, while a label of ``0'' indicates that $t_2$ is not entailed in $t_1$. The dataset includes both positive and negative instances. Data was constructed based on observations from LLMs prediction result on \data. Details can be seen in Table~\ref{tab:nli_data}.

Firstly, we used \textit{glm-4-9b} to convert the existing structural data into natural language expressions as positive instances, assigning all labels a value of ``1''.

We then designed three strategies for generating negative instances, namely ``Remove'', ``Revise'', and ``Replace''. By randomly selecting negative instance generation strategies to modify the certain proportion of the positive instances, we change the label information from ``1'' to ``0'' to obtain the negative instances.

We used the event extraction dataset CMNEE to construct data for training the NLI model for Event Recall and Argument Recall, translated the version of MAVEN-ERE for Causal Recall, automatic constructed data of \data for Temporal Recall.

\paragraph{Event Recall}

$t_2$ represents the sentence contained key event information, and the label is used to indicate whether the sentence is entailed in the input text $t_1$.

\begin{figure*}[t]
\centering
\includegraphics[width=1.0\linewidth]{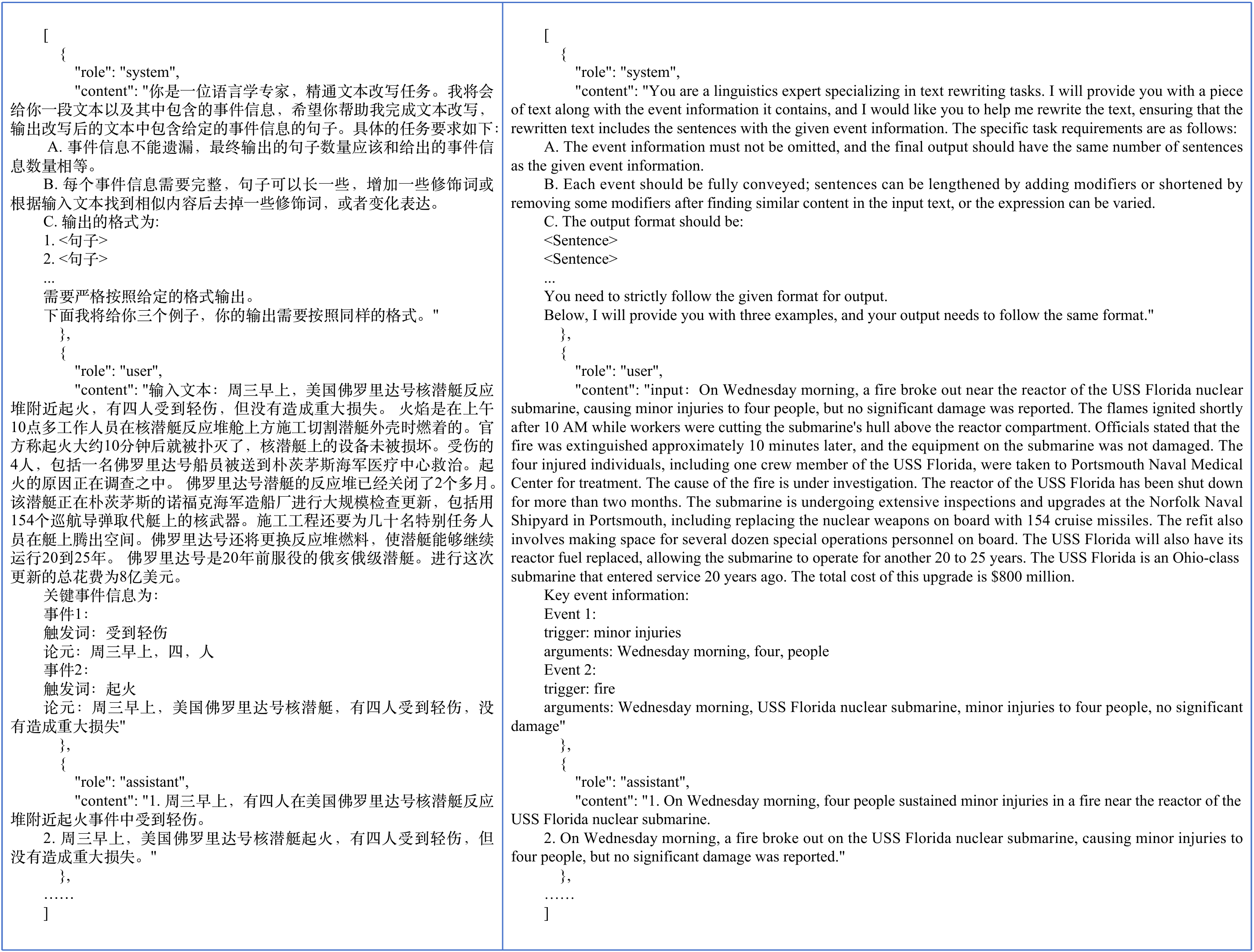}
\caption{Prompt used for LLMs to rephrase the structured event information into coherent natural language sentences to obtain positive instances of NLI models.}
\label{fig: positive_prompt}
\end{figure*}

To obtain positive instances, we used \textit{glm-4-9b} to rephrase the structured event information annotated in the dataset from the input text into a coherent natural language sentence. The prompt we used can be seen in Figure~\ref{fig: positive_prompt}. Then the input text is used as $t_1$, the generated key event information sentence is used as $t_2$, and the label is assigned a value of ``1''.

The modification methods corresponding to the negative instance generation strategies are as follows:

\begin{itemize}
\item Remove: Calculate the similarity between each sentence in the text and $t_2$, and remove sentences with similarity that exceed the certain threshold. The remaining low-similarity sentences were concatenated as the new text $t_1'$ and the label was set to ``0''.
\item Revise: Prompt the LLM \textit{glm-4-9b} to modify key information in the key event sentence $t_2$ from the positive instances, such as changing the ``time'', ``location'', ``quantities'', or ``person'' involved in the event. Alternatively, expand on or delete some information surrounding the key event in sentence $t_2$. The generated sentence after modification is used as the new sentence information $t_2'$ and the label was set to ``0''. Prompt can be seen in Figure~\ref{fig: revise_prompt}.

\begin{figure*}[t]
\centering
\includegraphics[width=1.0\linewidth]{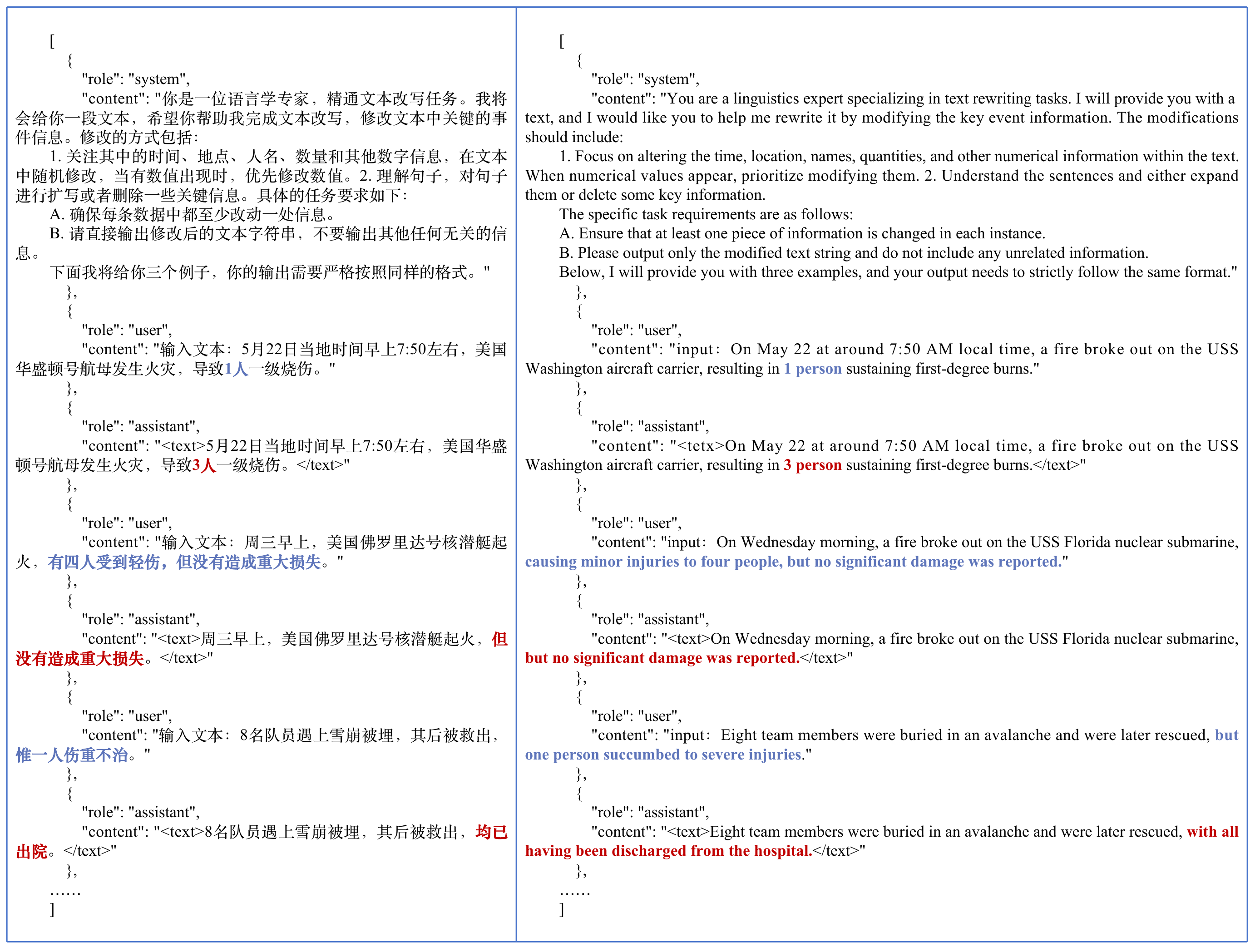}
\caption{Prompt used for LLMs to revise key event information to obtain negative instances of NLI models. The translated version is presented on the right of the figure. Relevant information before and after the revision was highlighted in blue and red respectively.}
\label{fig: revise_prompt}
\end{figure*}

\item Replace: Randomly retrieve 100 instances, calculate the similarity between the text of the retrieved instances and $t_2$, and use the text with the highest similarity but no overlapping event information as the new $t_1'$. The label was set to ``0''.
\end{itemize}

\paragraph{Argument Recall}
$t_2$ represents the key event argument, and the label is used to indicate whether the argument is entailed in the input text $t_1$.

Considering that arguments may include abbreviations, pronouns, etc., to more accurately evaluate argument recall, positive instances consists of two parts: 1) For ordinary arguments not in the coref-arguments list, use the text as $t_1$ and the argument information as $t_2$. 2) For arguments in the coref-arguments list, replace all other occurrences of the argument in the text with one of the arguments from the list. The modified text is used as $t_1$, and the argument information from the coref-arguments list is used as $t_2$. All labels are assigned as ``1''. Negative instance generated strategies were the same as those of Event Recall.

\paragraph{Causal Recall}

$t_2$ represents the sentence containing a causal relationship. The label is used to indicate whether the causal relationship is entailed in the input text $t_1$.

Currently, there is no existing Chinese dataset that meets our requirements for event causal relationships. Therefore, we translated the MAVEN-ERE dataset into Chinese using \textit{glm-4-9b} and obtained the input text after translation as $t1$ along with two sentences that exhibit a causal relationship. We then instructed the LLM to connect these two sentences into a coherent text using appropriate conjunctions, generating $t_2$. The labels were set to ``1''. Negative instance generated strategies were similar to those of Event Recall.

\paragraph{Temporal Recall}

$t_2$ represents the sentence containing temporal relationships, and the label indicates whether the temporal relationship is entailed in the input text $t_1$. The construction process was similar to that of Causal Recall.

\begin{table}[t!]
    \centering
{
    \begin{tabular}{lccc}
    \toprule
    {\textbf{Metric}} & {\textbf{Data Size}}\\
    \midrule
    Event Recall & $13,265/2,433/4,481$ \\
    Argument Recall & $15,000/3,000/3,000$\\
    Causal Recall & $10,082/3,505/4,098$ \\
    Temporal Recall & $9,678/1,461/1,318$ \\
    \bottomrule
    \end{tabular}
}
    \caption{Details of Data Constructed for NLI Models.}
    \label{tab:nli_data}
\end{table}

\subsection{Model Training}
We used \textit{sentence-transformers} to calculate text similarity and the relevant model was \textit{paraphrase-multilingual-mpnet-base-v2} good support for Chinese corpora and summarization task. During NLI model training, the max sequence length is set to 512, the label number is 2, and the training epochs are 6.

\section{C. Experimental Details}
\label{experimental details}
\begin{figure*}[t]
\centering
\includegraphics[width=1.0\linewidth]{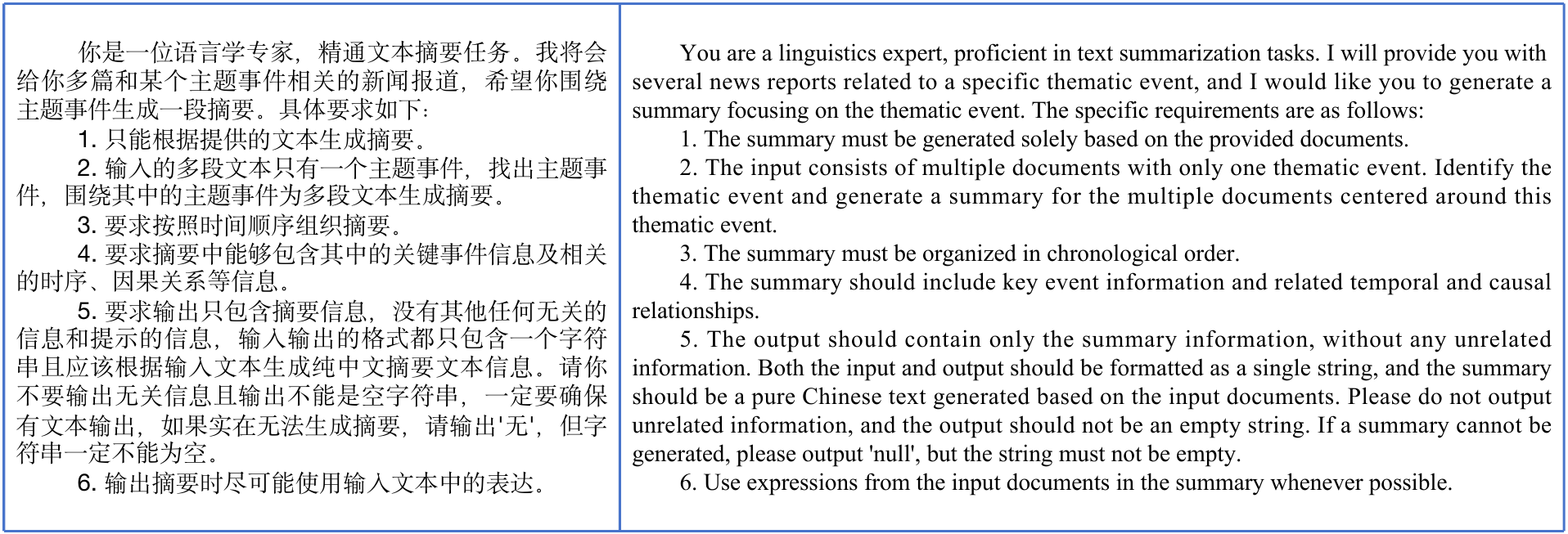}
\caption{Prompt for LLMs to generate summaries according to the input documents. The translated version is under the original prompt in the figure. }
\label{fig: sum_prompt}
\end{figure*}

The assessment of \data used popular LLMs was conducted under the zero-shot setting. The prompt can be seen in Figure~\ref{fig: sum_prompt}. We conducted all experiments on Nvidia A100 GPUs. All temperature parameters are set as 0.01.

\section{D. Few-shot Experimental Results}
In that section, we did preliminary few-shot in-context learning experiments, mainly testing the performance in 1-shot and 3-shot settings. Relevant demonstrations were randomly selected from the training set. Due to the length limitation, some models could not generate inference results properly. The detailed experimental results are shown in Table~\ref{tab:1-shot_result} and Table~\ref{tab:3-shot_result}, respectively. Considering the cost budget, we only measured open-source models results.

From the experimental results, it is evident that most of the metrics show a slight decline. Three reasons may account for this after observation:

1) The significant increase in input length poses a greater challenge to the long-context processing capabilities of LLMs, which has been validated in Figure~\ref{fig: futher_analysis}(a).

2) The provided demonstrations introduce a substantial amount of information unrelated to the target dynamic event, which poses a new challenge for the model to more carefully discern and extract valid information.

3) The summarized text exhibits greater stylistic diversity, and the inclusion of demonstrations partially constrains the style and content of the generated text.

We plan to conduct a more in-depth and detailed analysis in our future work.

\label{few-shot results}
\begin{table*}[t!]
    \centering
{
    \begin{tabular}{l|l|c|rrrr|rrrr}
    \toprule
    {\textbf{Type}} & {\textbf{Model}} & {\textbf{Length}}& {\textbf{R-1}} & {\textbf{R-2}} & {\textbf{R-L}} & {\textbf{BS}} & {\textbf{ER}} & {\textbf{AR}} & {\textbf{CR}} & {\textbf{TR}} \\
    \midrule
    \multirow{7}{*}{Open-Source} & 
      Baichuan2-13b-chat & $8$K & $-$ & $-$ & $-$ & $-$ & $-$ & $-$ & $-$& $-$ \\
    & Llama3-chinese-8B-Instruct & $8$K & $-$ & $-$ & $-$ & $-$ & $-$ & $-$ & $-$ & $-$\\
    & Yi-1.5-9b-chat-16k & $16$K & $29.9$ & $14.6$ & $19.1$ & $66.4$ & $14.5$ & $26.9$ & $42.7$ & $37.6$ \\
    & Qwen2-7b-Instruct & $32$K & $44.9$ & $24.9$ & $32.0$ & $75.0$ & $25.8$ & $41.0$ & $57.0$ & $42.6$ \\
    & glm-4-9b-chat & $128$K & $48.3$ & $27.3$ & $35.6$ & $77.2$ & $15.4$ & $30.4$ & $14.2$ & $15.8$\\
    & InterLM2.5-7B-Chat-1M & $1$M & $46.0$ & $26.1$ & $32.6$ & $75.4$ & $25.0$ & $42.6$ & $53.2$ & $40.5$\\
    & glm-4-9b-chat-1M & $1$M & $47.9$ & $27.9$ & $35.2$ & $76.6$ & $23.1$ & $40.9$ & $38.9$ & $30.0$\\
    \bottomrule
    \end{tabular}
    }
    \caption{1-shot experimental results. Length: Input length limitation of models; R-1: ROUGE-1; R-2: ROUGE-2; R-L: ROUGE-L; BS: BERTScore; ER: Event Recall; AR: Argument Recall; CR: Causal Recall; TR: Temporal Recall. Metric Definitions were illustrated in the Evaluation Metrics section. The best results are in bold. The second-best results are underlined.}
    \label{tab:1-shot_result}
\end{table*}

\begin{table*}[t!]
    \centering
{
    \begin{tabular}{l|l|c|rrrr|rrrr}
    \toprule
    {\textbf{Type}} & {\textbf{Model}} & {\textbf{Length}}& {\textbf{R-1}} & {\textbf{R-2}} & {\textbf{R-L}} & {\textbf{BS}} & {\textbf{ER}} & {\textbf{AR}} & {\textbf{CR}} & {\textbf{TR}} \\
    \midrule
    \multirow{7}{*}{Open-Source} & 
      Baichuan2-13b-chat & $8$K & $-$ & $-$ & $-$ & $-$ & $-$ & $-$ & $-$& $-$ \\
    & Llama3-chinese-8B-Instruct & $8$K & $-$ & $-$ & $-$ & $-$ & $-$ & $-$ & $-$ & $-$\\
    & Yi-1.5-9b-chat-16k & $16$K & $-$ & $-$ & $-$ & $-$ & $-$ & $-$ & $-$ & $-$ \\
    & Qwen2-7b-Instruct & $32$K & $42.6$ & $24.2$ & $30.9$ & $71.9$ & $21.8$ & $32.8$ & $37.5$ & $31.9$ \\
    & glm-4-9b-chat & $128$K & $47.7$ & $27.3$ & $35.5$ & $76.9$ & $16.1$ & $28.9$ & $15.8$ & $16.2$\\
    & InterLM2.5-7B-Chat-1M & $1$M & $47.1$ & $27.5$ & $33.6$ & $76.1$ & $26.4$ & $42.1$ & $52.6$ & $40.4$\\
    & glm-4-9b-chat-1M & $1$M & $47.6$ & $28.0$ & $35.4$ & $76.3$ & $22.1$ & $38.1$ & $38.7$ & $28.2$\\
    \bottomrule
    \end{tabular}
    }
    \caption{3-shot experimental results. Length: Input length limitation of models; R-1: ROUGE-1; R-2: ROUGE-2; R-L: ROUGE-L; BS: BERTScore; ER: Event Recall; AR: Argument Recall; CR: Causal Recall; TR: Temporal Recall. Metric Definitions were illustrated in the Evaluation Metrics section. The best results are in bold. The second-best results are underlined.}
    \label{tab:3-shot_result}
\end{table*}

\section{E. Cases for Generated Summaries Analysis}
\label{case analysis}
\begin{figure*}[t]
\centering
\includegraphics[width=1.0\linewidth]{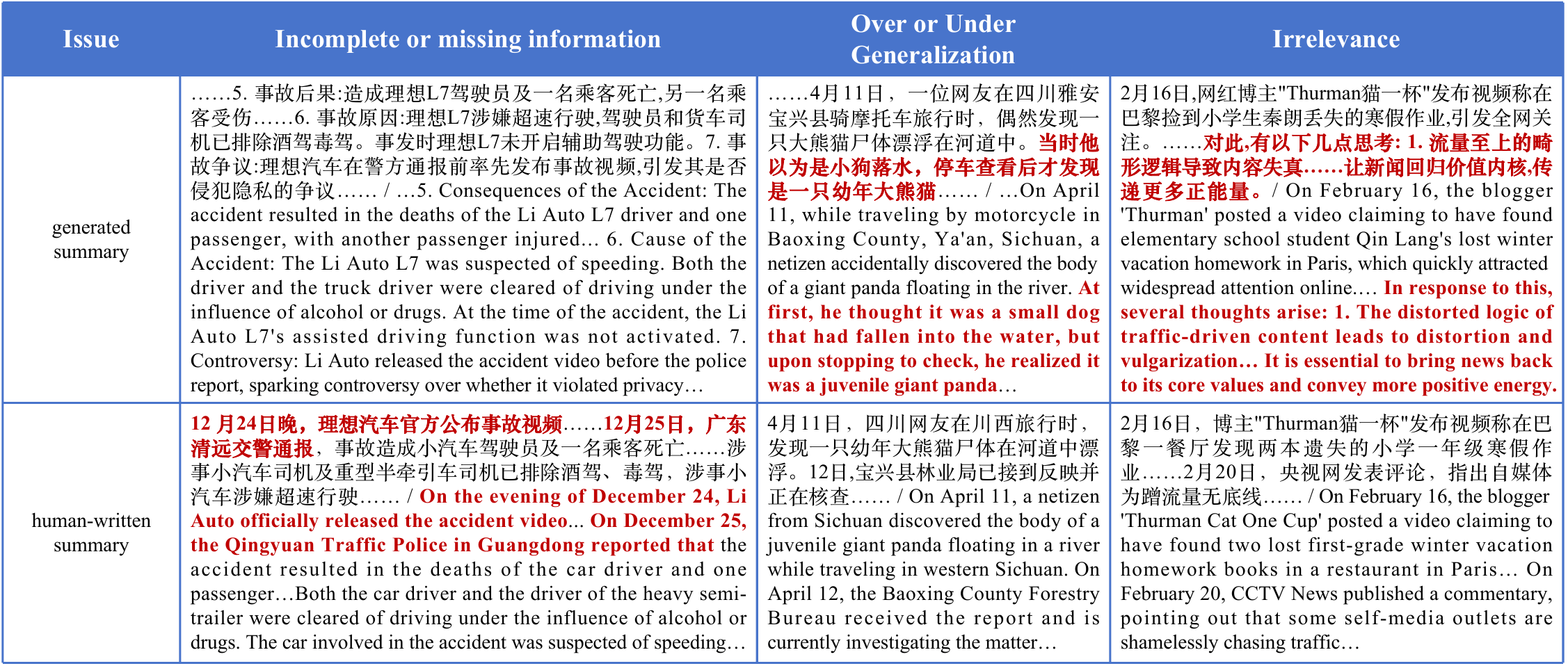}
\caption{Cases for generated summaries analysis. Each column represents a corresponding example of its specific type.}
\label{fig: case_study}
\end{figure*}

In this section, we provided a corresponding example for each issue, as illustrated in Figure~\ref{fig: case_study}. In Example 1, key time information and related sub-event information are missing. In Example 2, the generated summary is under-generalized and includes some detailed descriptive text which is not helpful in sorting out key event information. In Example 3, the generated summary contains some evaluative and reflective comments that are not directly related to the development of the event. We set the color of the problematic text to red in the figure.

\end{document}